# An Evolutionary Hierarchical Interval Type-2 Fuzzy Knowledge Representation System (EHIT2FKRS) for Travel Route Assignment

Mariam Zouari* *Student Member, IEEE*, Nesrine Baklouti* *Member, IEEE*,
Javier Sanchez Medina†, Mounir Ben Ayed* and Adel M. Alimi*, *Senior Member, IEEE*

*REGIM-Lab: REsearch Groups in Intelligent Machines, University of Sfax, National Engineering School of Sfax (ENIS), BP 1173, Sfax, 3038, Tunisia
†CICEI: Innovation Center for the Information Society University of Las Palmas de Gran Canaria Las Palmas de Gran Canaria, Spain

*Abstract*—Urban Traffic Networks are characterized by high dynamics of traffic flow and increased travel time, including waiting times. This leads to more complex road traffic management. The present research paper suggests an innovative advanced traffic management system based on Hierarchical Interval Type-2 Fuzzy Logic model optimized by the Particle Swarm Optimization (PSO) method. The aim of designing this system is to perform dynamic route assignment to relieve traffic congestion and limit the unexpected fluctuation effects on traffic flow. The suggested system is executed and simulated using SUMO, a well-known microscopic traffic simulator. For the present study, we have tested four large and heterogeneous metropolitan areas located in the cities of Sfax, Luxembourg, Bologna and Cologne. The experimental results proved the effectiveness of learning the Hierarchical Interval type-2 Fuzzy logic using real time particle swarm optimization technique PSO to accomplish multiobjective optimality regarding two criteria: number of vehicles that reach their destination and average travel time. The obtained results are encouraging, confirming the efficiency of the proposed system.

## I. Introduction

**D**YNAMIC Traffic Management is a very important component of human life and economy, which leads to positive effects on throughput, pollution and safety. The objective of road traffic management is not only to improve traffic fluency on road networks, assigning dynamically the traffic flows, but also to reduce the number of traffic congestions states and their negative effects (i.e. delays, waiting time, drivers stress, carbon footprint or emergency vehicles blocking) [1].

Dynamic Traffic Assignment is a key topic for the most advanced traffic management systems. Essentially, it pertains to the automatic diversion of the traffic into the ideal choice of a set of alternatives or the combination of alternatives towards each vehicle's destination node. A full-fledged dynamic traffic assignment system could, hypothetically, get along with a highly variable traffic user demand and conditions, even with unexpected situations, like accidents.

Thus, there is a need for a predictive system that offers each driver, the best choice at each intersection for every destination and present traffic state pair. However, conventional traffic management decisions are made driven simply by engineers experience and static and outdated traffic behavior assumptions, with the subsequent gap to an utmost efficient use of the traffic infrastructures. Moreover, the measurement of several decisions, objectives and constraints are commonly difficult to be realized by crisp values. Hard computing models cannot be treated effectively with the transport decision-makers ambiguities and uncertainties.

To deal with these problems in road perceptions, the application of Interval Type-2 Fuzzy Logic was considered an appropriate mathematical device for decision making in transportation engineering [2]. Actually, guaranteeing control by using Interval Type-2 Fuzzy Logic has been proven to be a great success in a large variety of applications [3][4][5]. Nevertheless, it is depending on the method used in the design process that the performance of a designed controller is likely to change. The definition of the appropriate parameters of membership functions values of a Fuzzy Logic System (FLS) in both antecedent and ensuing parts is difficult because their calculation is time-consuming. The major problem in designing an FLS lies in how to fix the suitable fitting membership values. In the literature, the attention has extensively been drawn to the learning of fuzzy systems. Among the proposed methods, three major evolutionary, similar and popular methods are the Genetic Algorithms (GA) [6][7], Ant Colony Optimization algo- rithms (ACO)[8] and Particle Swarm Optimization algorithms (PSO)[9]. Concerning the former, it is an evolutionary opti- mization technique which is quick, simple and likely to be used for looking for the best solution in wide search space. In fact, PSO has been proven to be effective in several applications. This pa- per proposes a Hierarchical Interval Type-2 Fuzzy Knowledge Representation System (HIT2FKRS) based on Particle Swarm Optimization for Travel Route Assignment. The membership functions of a Hierarchical Interval Type-2 Fuzzy controller are instantly tuned by means of PSO technique. The aim of the design of the suggested system was to resolve the travel route assignment problem in a dynamic way and ameliorate the complete road network quality, particularly in case of congestions and jams, considering real-time traffic information and drivers' travel time to attein their destinations. To corroborate our methodology, we have used traffic simulation. Actually, the open source microscopic road traffic simulator (SUMO) has been used [10]. We have also used the Open Street Map in the SUMO traffic simulator to get nearer to reality.

The rest of the paper is organized as follows. Section 2 overflies the use of fuzzy logic, Interval Type-2 Fuzzy Logic and PSO for the route choice problem. Section 3 presents the basic concepts of Particle Swarm Optimization and the theory of Interval type-2 fuzzy logic. The details of the proposed Evolutionary Hierarchical Interval Type-2 Fuzzy Knowledge Representation System for travel route assignment are displayed in the fourth section. Indeed, this section firstly describes the Evolutionary Hierarchical Interval Type-2- Fuzzy Logic System (EHIT2FLS) for route choice evaluation and its components and secondly presents the significant role of the well-known traffic simulator SUMO. Section 5 exhibits the conducted simulations using SUMO, and then the analysis of the obtained results to validate our approach. The paper ends with a conclusion and summary of our contribution and some perspectives for future research work.

## II. LITERATURE REVIEW OF ROAD TRAFFIC MANAGEMENT

Thanks to the ability of techniques and methodologies from artificial intelligence to intelligently solve the complex problems related to transportation systems [11], researchers have been interested in using them. This section summarizes the latest road traffic management models that rely on Fuzzy logic, Interval Type-2 Fuzzy Logic, Particle Swarm Optimization and simulator of urban mobility (SUMO).

### A. Fuzzy Logic-based road traffic management

Many classical methods have been suggested to primarily contend with the problem of choosing the route by means of the discrete choice models as Logit and Probit models, in the majority of cases [12][13]. Such techniques cannot adapt with ambiguities and uncertainties of the perceptions of environment, and thus they are con- sidered as an inefficiency gap to account for the dynamicity and complexity of transportation systems.

Hence, many researchers are resolving transportation problems by using soft computing, specifically Fuzzy Logic, as a useful tool for treating those uncertainties [14]. Among the well-known transportation related engineering problems are traffic light management, traffic assignment and road traffic management [15][16][17] can be mentioned. Table I presents some research works pertaining to fuzzy logic-based road traffic management. Despite the fact that Type-1 Fuzzy Logic System (T1FLS) is considered as one of the well-known types of FLS, it remains an unsatisfactory representation of the real-time traffic uncertainties. Furthermore, it permits to tackle a restricted uncertainty level, yet road traffic management problems often face several sources and elevated levels of uncertainty [18]. Therefore, it is crucial to integrate Type-2 Fuzzy Logic in road traffic management.

### B. Type-2 Fuzzy Logic-based Road traffic management

Type-2 fuzzy set [19] represents an extension of the concept of Type-1-Fuzzy set. A more elevated uncertainty level is tackled by making membership function three dimensional. The membership function is named FOU (Footprint Of Uncertainty). It is bordered by the upper and lower membership functions.

The structures of both T2FLS and T1FLS are too similar and the unique difference lies in adding a type-reducer component. The output processor generates a Type-1 Fuzzy set output (from the type-reducer) or a crisp output (from the defuzzifier) [20]. Indeed, Type-2 Fuzzy sets offer a more precise way to design uncertainty in a system. For instance, they can handle linguistic uncertainties by modeling information's vagueness and lack of reliability.

Several prominent works, summarized in Table II, have dealt with many problems in traffic management that have been solved applying T2FLS [2][21][22][23]. The majority of the developed models for itinerary selection depended only on two or three possible alternatives taking into account the restricted number of criteria to prevail over the rule-explosion problem. This is an enhancement for the present work to further continuing and relying on a Hierarchical Interval Type-2 Fuzzy Logic system for the itinerary evaluation that can take into consideration numerous selection criteria to select the best itinerary.

### C. Swarm intelligence (PSO)-based road traffic management

The extensive use of swarm intelligence was to model complex traffic and transportation processes [24]. In fact, Ant Colony Optimization (ACO) has been extensively used to solve transportation problems, such as Vehicle Routing Problem (VRP) and Trav- eling Salesman Problem (TSP) [25][26]. However, there exist a few publications based on swarm intelligence, especially PSO, to solve road traffic management problems. PSO algorithm is a population intelligence algorithm that has good performance in optimization. The most important features of the pertinent related works on traffic management based on PSO are summarized in Table 3. In most of the presented studies, the PSO was applied to control and optimize traffic signals. This is the reason behind taking the initiative to test the reliability of PSO for learning our HIT2FLS to travel route assignment.

## III. FUNDAMENTALS OF PARTICLE SWARM OPTIMIZATION AND INTERVAL TYPE-2 FUZZY LOGIC

This section introduces the basic concepts of PSO and theory of the Interval Type-2 Fuzzy logic which is a generation of type-1 Fuzzy Logic.

### A. Particle Swarm Optimization (PSO)

Particle Swarm Optimization (PSO) is a simple and robust technique of optimization, which was inspired by bird flocking's social behavior. It was initially introduced by Eberhart and Kennedy in 1995 [9]. It looks like evolutionary techniques as Genetic Algorithms (GA). PSO operates with a

TABLE I
RELEVANT RELATED WORKS ON TRAFFIC MANAGEMENT BASED ON TYPE-1 FUZZY LOGIC

| Refs | Year | Objective | Main features | Strength |
|------|------|-----------|---------------|----------|
| [15] | 2014 | Traffic congestion prediction system | Hierarchical fuzzy system optimized by genetic algorithms | Decreases the input variables' size while preserving a good precision |
| [16] | 2014 | Road Traffic management | - Hierarchical fuzzy model - Multi-agent system based on the ant colony behavior | Fine-tune the road traffic in the network in line with the real times using vehicle route guidance system |
| [17] | 2016 | Congested road notification system | Fuzzy logic inVANET context aware congested road | Best Evaluation of the present vehicle situation to the congestion situation. |

TABLE II
RELEVANT RELATED WORKS ON TRAFFIC MANAGEMENT BASED ON TYPE-2 FUZZY LOGIC

| Refs | Year | Objective | Main features | Strength |
|------|------|-----------|---------------|----------|
| [2]  | 2011 | Urban traffic management | Multi-agent architecture based on type-2 fuzzy logic | - Boost the greentime - Diminish the total delay of vehicles - Real traffic simulation |
| [21] | 2012 | Traffic light control | Distributed multi-agent system based on type-2 fuzzy logic system for traffic signal control | - Define the green time time that diminishes the global delay |
| [22] | 2014 | Traffic congestion | Genetic algorithm (GA) for enhancing the membership function parameters of Type-2 fuzzy logic | - Decrease the queue length and vehicular delay at the intersection. |
| [23] | 2016 | Traffic flow data forecasting | Interval type-2 fuzzy sets theory for traffic flow data forecasting | - Lessen the impact of noise from the detection data. |

TABLE III
RELEVANT RELATED WORKS ON TRAFFIC MANAGEMENT BASED ON PSO

| Refs | Year | Objective | Main features | Strength |
|------|------|-----------|---------------|----------|
| [27] | 2012 | Traffic Signal Control for signalizing Roundabouts | - PSO - Fuzzy Logic | * Instantly respond to the current traffic condition of the roundabout in order to ameliorate real timeness. |
| [28] | 2013 | Traffic light control | PSO for finding efficient traffic light cycle programs coupled with the traffic simulator SUMO (Simulator of Urban MObility) | * Useful traffic light cycle programs for both cities : Bahia Blanca in Argentina and Malaga in Spain *Quantitative improvements for the two main objectives: the number of vehicles that reach their destination and the overall journey time. |
| [29] | 2013 | Traffic Signal Control | * The Bayesian Network (BN) model and the Cellular Automaton (CA) model are used to build up a probability model for traffic jam. *PSO based on the probability model | * Obtain optimal traffic signals * minimize the probability of traffic jam |
| [30] | 2016 | Traffic Signal Control | * Cell Transmission Model (CTM) so as to model the online traffic network and PSO algorithm to optimize the control of the traffic signal network | * Minimize the overall delay on the traffic network * Lower the fuel consumption. |
| [31] | 2017 | Traffic management and traffic light control | * Multi-Objective Particle Swarm Optimization (MOPSO) * Multi-agent system (MAS) * SUMO | * Novel traffic management model optimizes vehicle re-routing * Traffic light control to alleviate traffic congestion and limit the effects of incidents on traffic flow based on (MOPSO) method. |

---

**Algorithm 1: Pseudocode of PSO**

1. Initialize the swarm : position, velocity, the swarm size and constant C1 and C2.
2. Calculate the fitness value for each particle. Assess the particle's best position and the global best position of the swarm. Then, update the particle's velocity and position using equation (1).
3. If one of the stopping criteria is reached, then stop, else go to step 2.
4. The optimal solution is the one that offers the latest gbest.

---

solutions population called particles and the whole population is called the swarm. The swarm is initialized with random solutions. A swarm is composed of N particles in a M dimensional search space. The values of position and velocity are denoted for each particle of the swarm. Each particle consecutively adjust its position on the basis of two factors: the best position of its neighbors $p_{best}$ and that visited by the whole swarm $g_{best}$. The determination of the best solution of local or global swarm is realized by the fitness function that represents the PSO algorithm optimization criteria. In what follows the equations representing the velocity and the position of a $i^{th}$ particle respectively :

$$V_i(t+1) = w * V_i(t) + C_1 * R_1 * (p_{best} - X_i(t)) + C_2 * R_2 * (g_{best} - X_i(t)) \quad (1)$$

$$X_i(t+1) = X_i(t) + (1-w) V_i(t+1) \quad (2)$$

where $C_1$ and $C_2$ are constant factors, $w$ is the inertia weight. $R_1$ and $R_2$ are random numbers in [0, 1]. Algorithm 1 describes the pseudo-code of PSO.

Actually, the Particle Swarm Optimization algorithm is used in several engineering problems [32][33][34] indicating a successful performance, even in comparison with other contemporary optimization techniques [35]. Despite that, the use of PSO is still limited for travel route assignment and other
problems related to the road traffic management.

*B. Interval Type-2 Fuzzy Logic*

Interval Type-2 Fuzzy Sets (IT2 FSs), originally introduced by Zadeh [36], provide additional degrees of freedom in Mamdani and TSK Fuzzy Logic Systems (FLSs), that can be efficient in situations where lots of uncertainties and imprecision are present[37]. The Interval Type-2 Fuzzy Logic Systems (IT2FLS) have the potential to afford better performance than a Type-1 Fuzzy Logic System. As a result of the uncertainties present in the surrounding environments, Type-1FLS might not be suitable because they may cause deterioration in the FLS performance. We might finish wasting time in frequently tuning or redesigning the Type-1 FLS so that it can deal with the various uncertainties encountered. Type-2 FLSs employing Type-2 Fuzzy sets can handle such high levels of uncertainties to offer very good performances and provide better modeling. A Type-2 fuzzy set is characterized by a fuzzy membership function, so the membership value for each element of this set is a fuzzy set in [0,1], unlike a type-1 fuzzy set in which the membership value is a crisp number in [0,1]. Unlike fuzzy logic, which uses two-dimensional memberships, Type-2 fuzzy logic is characterized by three dimensional memberships. For Type- 2 fuzzy logic the membership function is called FOU (Footprint Of Uncertainty) and it is delimited by the lower and the upper memberships (LMF and UMF respectively). The new third dimension provides additional degrees of freedom to model and handle uncertainties [38]. A Type-2 FLS is very similar in structure to a Type-1 FLS, the only difference is the extra output process component which is called the type-reducer before defuzzification, as we can be seen in Figure 1 for a Mamdani model [39].

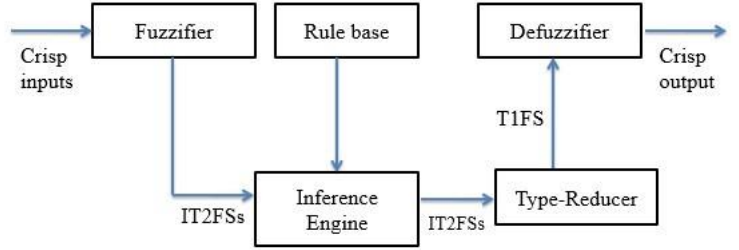

Fig. 1. The structure of a Type-2-FLS

In details, an IT2 FLS works as follows : the crisp inputs are first fuzzified to input Type-2 fuzzy sets that are fed to the inference engine which maps the input Type-2 Fuzzy sets to output Type-2 Fuzzy sets using the rule base. The output set is then processed by the type-reducer in the type reduction section that generates a Type-1 output set.

Recall that a general Type-2 fuzzy set, named $A$, is expressed in the universe of discourse X by a Type-2 membership function $\mu_{\widetilde{A}}(x, u)$ :

$$\widetilde{A} = \{((x, u), \mu_{\widetilde{A}}(x, u)) | \forall x \in X, \forall u \in J_x \subseteq [0, 1]\} \quad (3)$$

with $0 < \mu_{\widetilde{A}}(x, u) < 1$

$$\widetilde{A} = \int_{x \in X} \int_{u \in J} \mu_{\widetilde{A}}(x, u) / (x, u) J_x \in [0, 1] \quad (4)$$

with $\int\int$ design the union of all admissible $x$ in $u$. For an Interval Type-2 Fuzzy Set (IT-2FS), the membership function $\mu_{\widetilde{A}}(x, u)$ is equal to 1. Equation 4 becomes then:

$$\widetilde{A} = \int_{x \in X} \int_{u \in J} 1/(x, u) J_x \in [0, 1] \quad (5)$$

$LMF$ and $UMF$ are expressed by $\underline{\mu}_{\widetilde{A}}(x)$ and $\overline{\mu}_{\widetilde{A}}(x)$ respectively. So, the $FOU$ is defined by the following relation :

$$FOU(\widetilde{A}) = \bigcup_{x \in X} u \in J_x; J_x = \left[\overline{\mu}_{\widetilde{A}}(x), \underline{\mu}_{\widetilde{A}}(x)\right] \quad (6)$$

## IV. ADAPTIVE FUZZY LOGIC CONTROL BY PSO OPTIMIZATION

In order to achieve the desired level of robust performance for a controller, the exact tuning of the membership functions is very important. This paper intends to apply PSO algorithm in order to dynamically adjust the MFs of our HIT2FKRS model and design optimal fuzzy logic controllers (FLC). The approach of using a PSO for control rules and their corresponding MFs tuning in FLSs is shown in Figure 2. The proposed real time PSO algorithm is expressed by the following sequential steps that are executed at each time step:

- Step1: A swarm of M particles or N Fuzzy systems is initialized. It becomes an initial swarm. At the same time, the initial values of the velocity vectors or also noted variation of all particles' positions are randomly generated in the range $[-V_{max}, V_{max}]$. Each FLC in the initial population is assessed using the fitness function. Fetch the best value of the fitness function pbest.
- Step2: Variation updating using equation (7) while checking the maximum velocity.
- Step3: Using the updated velocities, each particle changes its position according to the following equations:

$$v_{imf}(k+l)(t) = w \times v_{imf}(k)(t) + c1 \times rand() \\ \times (pbest_{imf}(k)(t) - x_{imf}(t)) + c2 \times rand() \\ \times (gbest(k)(t) - x_{imf}(t)) \quad (7)$$

$$x_{imf}(k+1)(t) = x_{imf}(k)(t) + v_{imf}(k)(t) \quad (8)$$

where k is the $k^{th}$ iteration, $t$ is the time step, $w$ is the inertia weight, $rand()$ indicates a random function in [0 1], $pbest_i$ denotes the local best of the $i^{th}$ particle, gbest indicates the global best of the swarm and ($c1$, $c2$) are cognitive and social parameters.
- Step4: Update of pbest and gbest.
- Step5: Stopping criteria. If one of the stopping criteria is achieved, then stop, or else go to step 2.
- Step6: The FLC generating the latest gbest is the optimal FLC.

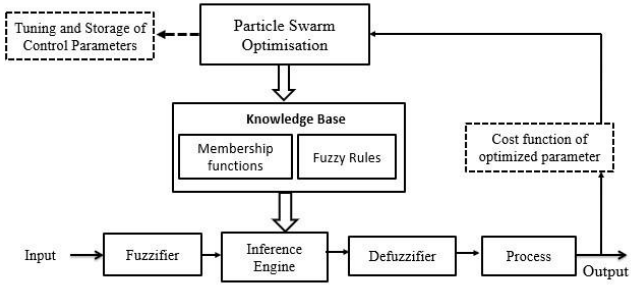

Fig. 2. The scheme of tuning of the fuzzy controller using PSO

The PSO algorithm trains at each step time the FLC parameters to extract instantaneously the best values that help the drivers achieve his/her destination, while avoiding congestions. The PSO-FLS process works as described in Figure 3.
In this process, each particle is shaped to represent the fuzzy control rules and their corresponding MFs parameters of the FLCs inputs and outputs. Each particle represents a potential solution. These parameters are used to define the particles of PSO algorithm and look for the global best fitness.

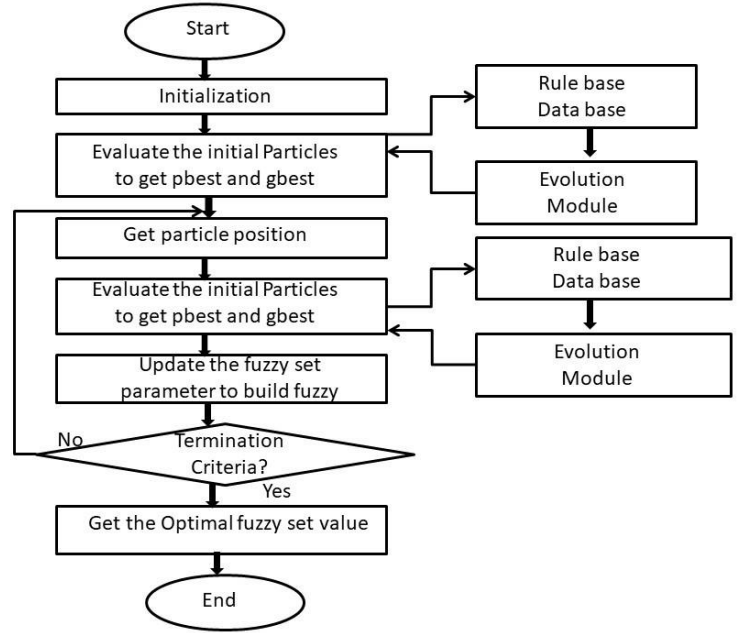

Fig. 3. Flowchart of Fuzzy Membership Function adjusting using PSO

## V. ARCHITECTURE OF OUR EHIT2FKRS : EVOLUTIONARY HIERARCHICAL INTERVAL TYPE-2 FUZZY KNOWLEDGE REPRESENTATION SYSTEM

This section presents in details the proposed Evolutionary Hierarchical Interval Type-2 Fuzzy Knowledge Representation System (EHIT2FKRS) for travel route assignment.
The objective of the suggested system is to select the best itinerary for each vehicle in terms of the real-time road traffic quality by the addition of other factors associated with the decision-making process, namely environmental conditions, infrastructure and driver characteristics. Thus, the suggested EHIT2FKRS for travel route assignment involves two main phases. Concerning the first one, it is initiated from the road-traffic simulator SUMO. As for the second phase, it is based on an EHIT2FL model to select the best itinerary from the set of possible itineraries by providing information from the simulator SUMO that has an influence on the route choice.

### A. Evolutionary Hierarchical Interval Type-2 Fuzzy (EHIT2F) model for travel route assignment

To arrive at destination, the proposed system increases the average speed of vehicles on the road network, while choosing the best itinerary according to the real-time road traffic quality. It is noteworthy that the best route is selected based not only on the distance and the expected travel time, but also on contextual factors linked to the driver, the environment, and infrastructure. Because of the ambiguity, uncertainty and dynamicity of these factors, it is very difficult to formulate a suitable deterministic mathematical model. However, the development of Interval type-2 fuzzy logic seems justified in this situation. With respect to the growing number of selection criteria used to choose the best alternative, the application of fuzzy logic to route choice problem with a large number of inputs causes the problem of rule-explosion. The use of hierarchical architecture seems to be the good solution to surmount this problem. As regards the traffic management, this tool was used effectively in some research works [16][40][41]. The present work develops a HIT2FLS for itinerary evaluation, which is fed with six fuzzy inputs that influence route choice decision: density, maximum speed allowed in the itinerary, familiarity of the driver with the path i, usual driving speed, departure time and weather information. The structure of HIT2FLS for itinerary evaluation is presented in Figure 4. In some fuzzy hierarchical architectures, outputs can be the inputs of the following fuzzy layer [40], compromising the interpretability of the overall model, since the intermediate outputs do not have a physical meaning.

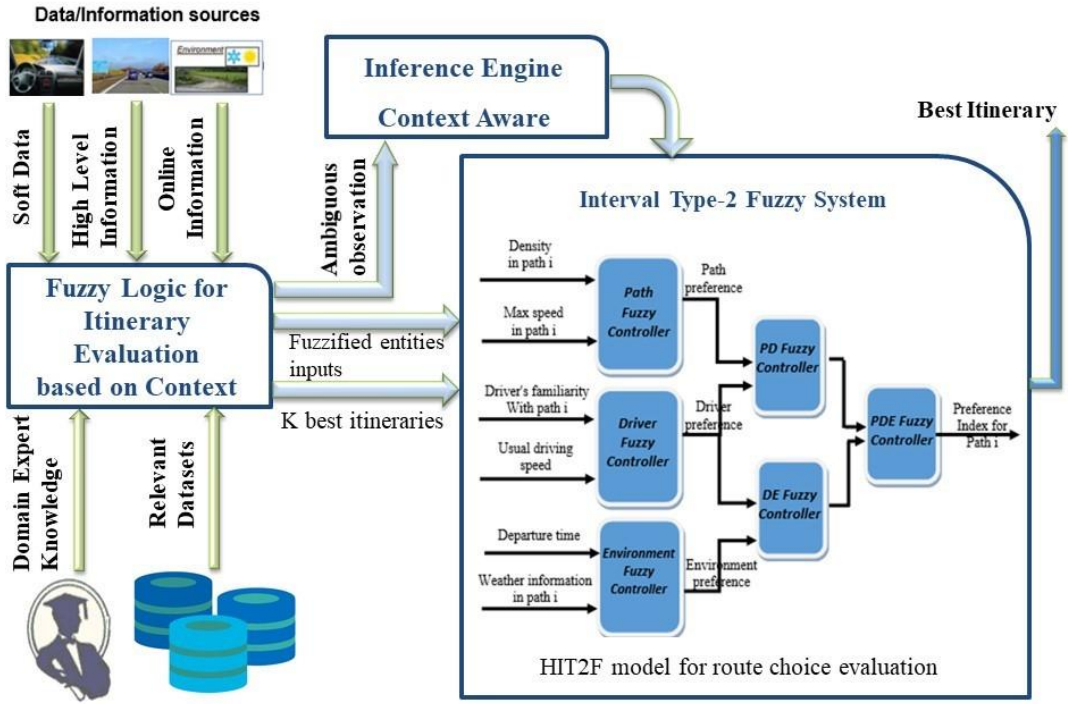

Fig. 4. The architecture of our proposed approach

To overcome this problem, the combination of inputs was considered as a resort to reduce limitations linked to the physical meaning loss in intermediate outputs/inputs. Indeed, inputs are reorganized in three categories, namely the itinerary criteria, the driver criteria and the environment criteria.

Our HIT2FLS has 6 fuzzy controllers, each one of them has two inputs and one output. Figure 5 illustrates the fuzzy sets of each input. The inference process of the fuzzy controllers is the Mamdani(max-min) inference method. Table 4 presents the fuzzy rule base of the path fuzzy controller, based on the ordinary rules of the type IF condition THEN action. Further details about these fuzzy controllers are presented in a previous research works [42].

In the majority of fuzzy logic controller applications, the shapes of membership functions are always chosen by humans arbitrarily. However, it cannot ensure providing the optimal control for the corresponding system. To promote the performance of the fuzzy controller, a PSO algorithm is adopted to optimize the shapes of the membership functions. The PSO algorithm trains at each step time the FLC parameters of the final block (PDE fuzzy controller) since we do not have an interpretable suitable fitness function for the other type-2 fuzzy blocks. Therefore, the purpose of PSO algorithm application in fine-tuning the membership functions of PDE fuzzy controller is to further help the driver to be well guided to his/her destination, while circumventing jam/congestion situations. Hence, in order to find the candidate solution that has the best performance, the positions of all the particles should be evaluated by fitness function.

TABLE IV
FUZZY RULE BASE OF PATH PREFERENCE.

| Rule no | Inputs | | Output |
|---|---|---|---|
| | Density of path i | Maximum speed in path i | Path preference |
| 1 | synchronized | Free | Medium |
| 2 | synchronized | synchronized | Weak |
| 3 | synchronized | Jam | Weak |
| 4 | Free | synchronized | Strong |
| 5 | Free | Jam | Medium |
| 6 | Jam | synchronized | Weak |
| 7 | Jam | Free | Weak |
| 8 | Jam | Jam | Weak |
| 9 | Free | Free | Strong |

The fitness function of our algorithm depended principally on a central criterion which is the density in the path. Our objective was to minimize the fitness function calculated using the following equation:

$$fitness = density\ of\ path\ i\ /\ Total\ density\ of\ possible\ path \quad (9)$$

$$density = (vehicle\ count\ /\ length\ edge) \quad (10)$$

Several experiments have allowed us to discover that setting c1 and c2 equal to respectively to 2 gets the best general performance. Nonetheless, regarding the inertia parameter w, it was chosen to be fixed to a value equal to 0,99 according to many tests as well.

To assess the proposed real-time algorithm, we chose to evaluate two different sized population swarms corresponding to population size of 30, 80. We set the maximum number of iterations to NI = 10 iterations.

*B. Road traffic simulations*

Traffic simulation is an important tool for modeling the operations of dynamic traffic systems. It helps analyzing the causes and potential solutions of traffic problems such as traffic jams, congestion and traffic safety. We are interested in the microscopic type of simulation. There is a lot of traffic simulators available nowadays with different features, such as VISUM, Vissim, CORSIM, MATsim and SUMO.

In our research study, the simulator SUMO was chosen as it enables its users to load different road networks and set various traffic streams. The Simulation of Urban Mobility (SUMO) helps to investigate several research topics like Route choice, traffic light and communication simulation between vehicles [43]. It is thanks to its numerous benefits that SUMO is extensively used. Firstly, it is an open portable source for microscopic road traffic simulation [44]. Furthermore, this package has the capacity to design both road network infrastructure and traffic demand.

A network (.net.xml file) comprises the information about the map structure: nodes (junctions), edges (streets), and the connections between them. The network can be imported from OpenStreetMap (OSM) [45] which is a widespread digital map. A trip is a vehicle movement from one location to another defined by: the starting edge, the destination edge, and the departure time.

A route is an extended journey, i.e., a route definition contains the first and the last edges, along with all the edges the vehicle will pass through. These routes are stored in a demand file (.rou.xml file). Supplementary files (.add.xml) can be added to SUMO information pertaining to the map or the traffic lights. SUMO permits the replacement and edition of information on the cycles of traffic lights by manipulating a file with .add.xml extension. The output of a SUMO simulation is recorded in a journey information file (.tripinfo.xml) that contains information relating to each vehicles departure time, the time the vehicle waits to start at (offset), the time the vehicle arrives, the duration of its journey, and the number of steps in which the vehicle speed is below 0.1m/s (temporal stops in driving). Thanks to the Traffic Control Interface (TraCI)[46], SUMO provides a high level of flexibility such that the user can retrieve, change the objects in the simulation and adapt the simulation online.

Due to TraCI, SUMO can be operated through script programming, both in Python (TraCI-Python) in Java (TraCI4j) or in Matlab (TraCI4Matlab)[47]. TraCI takes into account the real-time environment and constraints in the simulation.

TABLE V
ROUTING ALGORITHM

| Routing algorithm |
|---|
| 1. Initialize the trip source and destination |
| 2. If car i detect an intersection j |
| 3. Search a set of the following possible roads |
| 4. Assess each subsequent possible road using evolutionary HIT2FLS |
| 5. Compare the preference index of each path and select the best itinerary having the maximum preference index value |

VI. PERFORMANCE EVALUATION OF EHIT2FKRS

The traffic simulations is the best recourse when the traffic management requires a clear comprehension of the flows, especially jam or congestion situations. Indeed, they demonstrate their ability to predict efficient solutions to complex problems like routing problem[27][31][48][49].

The routing problem in a road network is to discover strategies to assign routes to vehicles in order to minimize their travel time and reduce congestion on the network. Thus, the route assignment or route choice process is significant to enhance the traffic fluency situation and road quality. It pertains to the selection of alternative set of routes between origin and destination in road networks. The simplest strategy found in literature is Dijkstra's algorithm [50], in which each vehicle takes the shortest path between origin and destination. Although all modern routing algorithms allow tremendously fast calculation of the shortest paths, they necessitate some preprocessing of the traffic network and storage of data that is produced from this preprocessing.

Therefore, this paper introduces a route assignment algorithm using not only the shortest path information but also some knowledge of congestion on different roads. The suggested EHIT2FKRS is carried out before each intersection to select the best following road to attain destination evading crowded areas, and thus conceding time and route length.

The results of the proposed approach were compared with those of the well-known Dijkstra's algorithm, HIT2FKRS, HIT1FKRS and those of HIT1FKRS optimized by PSO based on two criteria which are the average travel time and path flow. Table 5 summarizes the routing algorithm.

In this study, our proposal was tested with four large and heterogeneous metropolitan areas with hundreds of vehicles and traffic lights located in the cities of Sfax (Tayeb Mhiri zone), Luxembourg (LuST) in French, Bologna and Cologne in Germany. The characteristics of the selected realistic traffic areas in the present paper are as follows:

Scenario 1:
We imported the map of Tayeb Mhiri zone (Sfax, Tunisia) from OpenStreetMap (OSM) (Figure 6). In particular, we edited the map using JOSM (Java OpenStreetMap Editor) [45] before applying SUMOs Netconvert. JOSM is used to extract and manually select and change points of interest and road segments. The scenario covers an almost 18 $km^2$ area.
In the SUMO network file, we define an edge as a segment between two nodes, which can be divided into one or more lanes. Once a road network is imported, traffic demand and routes for each vehicle should be created. Two likely applications, namely DUAROUTER and DFROUTER, exist to compute routes. It is thanks to these routing applications that a route file containing the routing information for each vehicle defined in the network was produced. Table 6 reveals information about the chosen topology for the road network.

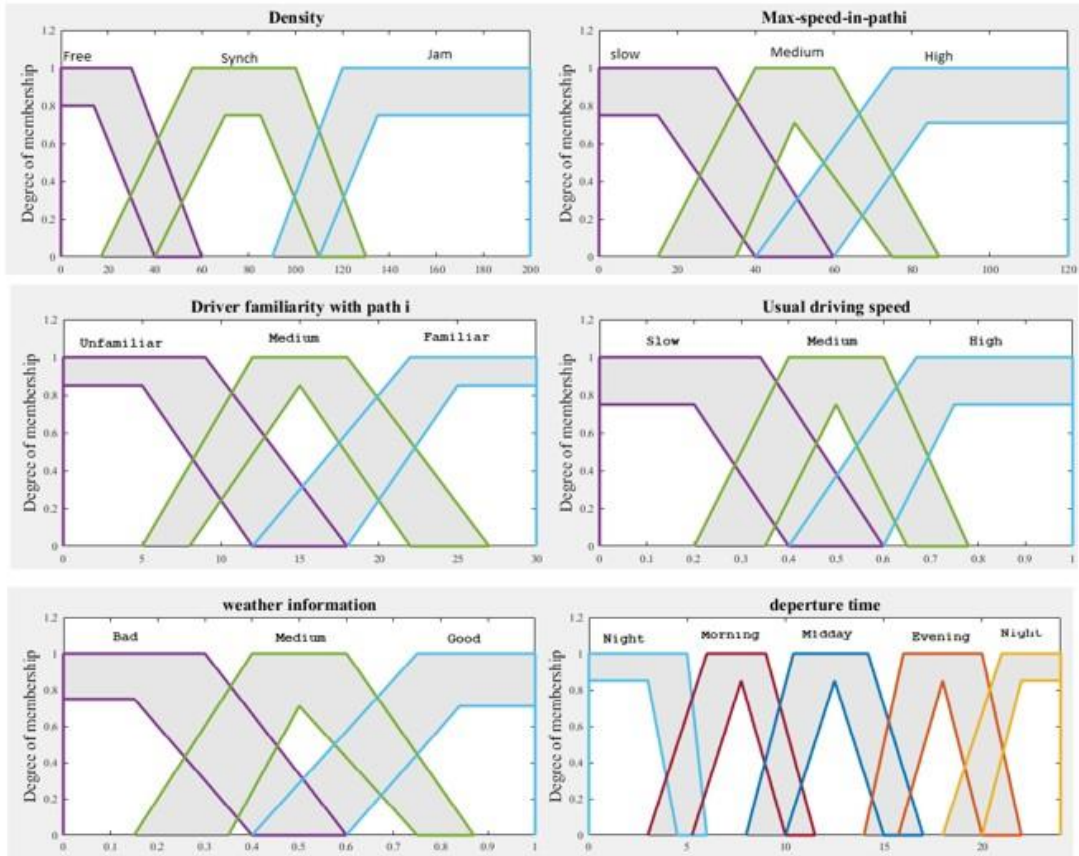

Fig. 5. Inputs membership functions

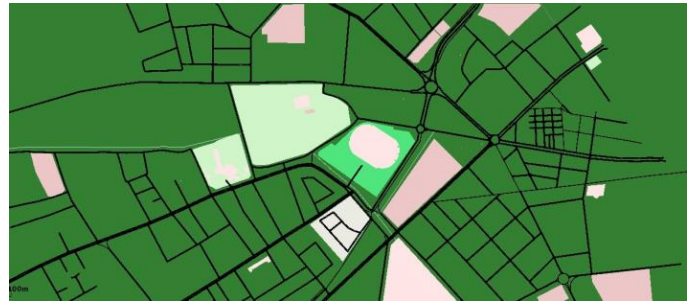

Fig. 6. SUMO net file of Tayeb mhiri zone

Scenario 2:
To have a realistic scenario, the area should be big enough to display the congestion patterns observable in contemporary cities. Thus, we chose the City of Luxembourg that covers an area of almost 156 $km^2$ with a total of 931 km of roads. There are more than 4,000 intersections, with 200 of them regulated by the traffic signal system. The city's public transport database was used to retrieve information about bus routes.

A total of 561 bus stops were introduced in the scenario, with 38 bus routes inside the city for a total of 2,240 buses over the 24 hour period [51]. Figure 7 shows the topology of the LuST Scenario. This version and the former versions of the LuST Scenario are used by different research groups and studies to assess ITSs and VANET protocols [52] [53]. The static information contained in the roads network topology file is summarized in Table 6.

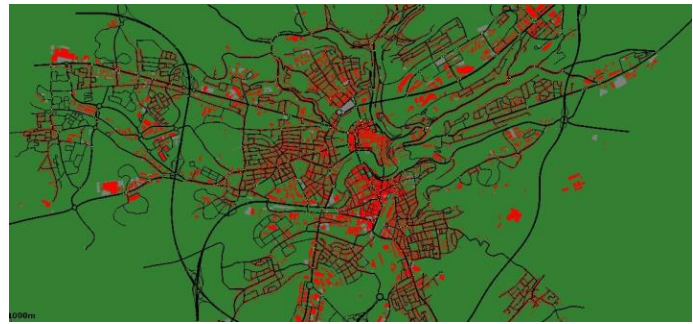

Fig. 7. LuST Scenario Topology

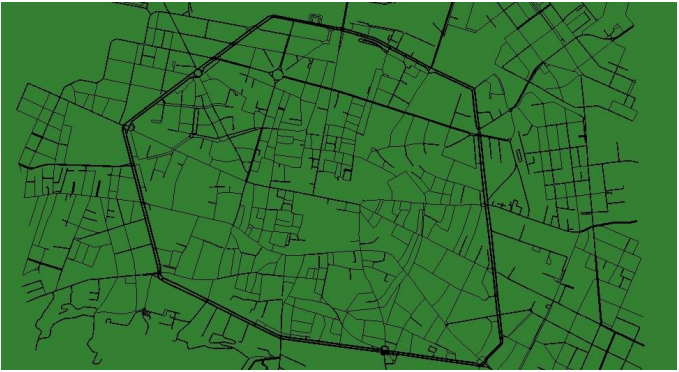

Fig. 8. Bologna Scenario Topology

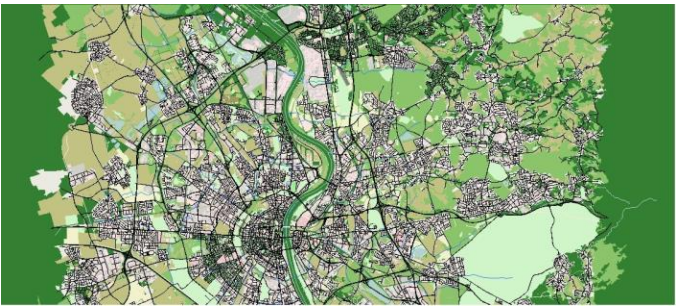

Fig. 9. TAPASCologne Scenario Topology

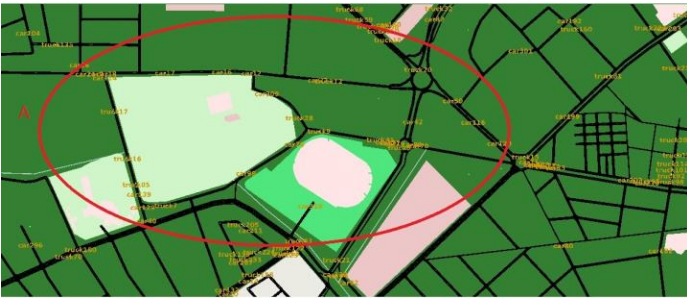

Fig. 10. Example of simulated road network

TABLE VI
TOPOLOGY INFORMATION

|  | Scenario 1 | Scenario 2 | Scenario 3 | Scenario 4 |
|---|---|---|---|---|
| Area ($km^2$) | 18 | 156 | 25 | 400 |
| Total intersections | 362 | 2,372 | 1539 | 31614 |
| Total roads | 852 | 5,969 | 2856 | 71085 |
| Traffic lights | 15 | 203 | 99 | 1219 |
| Number of vehicles | 500 | 2,336 | 22000 | 700,000 |

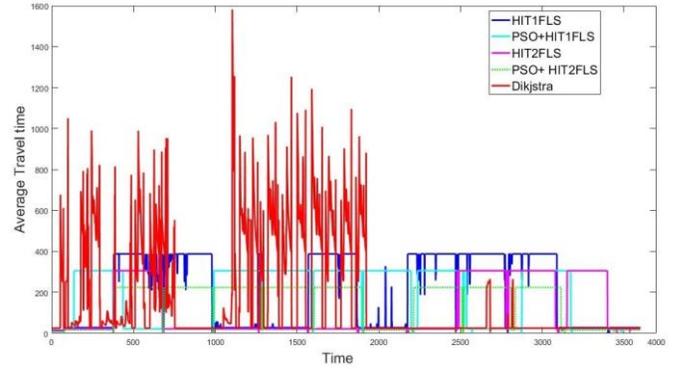

Fig. 11. Average travel time of cars (from origin to destination)(scenario 1)

Scenario 3:
Bologna Ringway Dataset models a realistic traffic scenario from the city of Bologna [54], Italy, during a typical day between 8AM and 9AM with more than 22000 vehicles in a 25 $km^2$ area. Figure 8 reveals the topology of the Bologna Scenario. The traffic demand is defined only over one hour in a typical morning. The data was produced beginning from Induction Loops, amassed by the iTETRIS European collaborative project [55][56]. A total of 22,213 individual trips is present in the dataset, beginning at 93 different edges, and ending at 81 edges around the city. In this case the scenario covers only the main streets. Table 6 summarizes the static information found in the roads network topology file.

Scenario 4:
The "TAPAS Cologne" simulation scenario describes the traffic within the city of Cologne (Germany) for a whole day [57]. The demand data emanates from TAPAS, which is a system that computes mobility wishes for a generated area population. The latter is based on information pertaining to the traveling habits of Germans and the infrastructure of the area they live in. The traffic demand information on the macroscopic traffic flows across the Cologne urban area (the O/D matrix) is obtained through the Travel and Activity PAtterns Simulation (TAPAS) methodology. The traffic assignment of the vehicular flows described by the TAPASCologne O/D matrix over the road topology is achieved using Gawrons dynamic user assignment algorithm. The city of Cologne covers a region of 400 square kilometers for a period of 24 hours in a typical working day, and comprises more than 700.000 individual car trips.

### A. Results and discussion

Series of simulation are carried out when changing the travel demand level, origins and destinations, departure time and congestion/jam position. The results of the proposed method were compared to those of static method based on the itinerary length using the algorithm of Dijkstra (1959) to the results of HIT2FKRS, HIT1FKRS and the results of HIT1FKRS optimized by PSO.

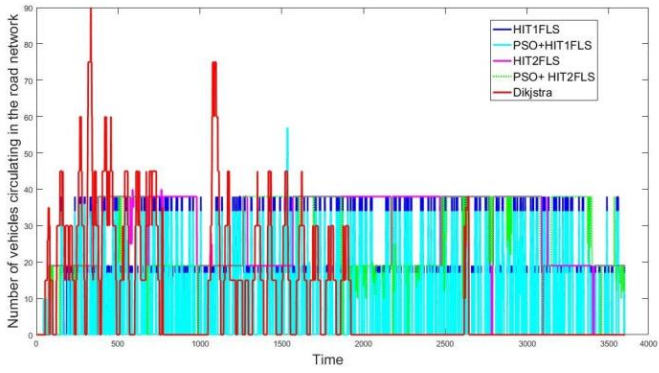

Fig. 12. Number of vehicles circulating in the road network (scenario1)

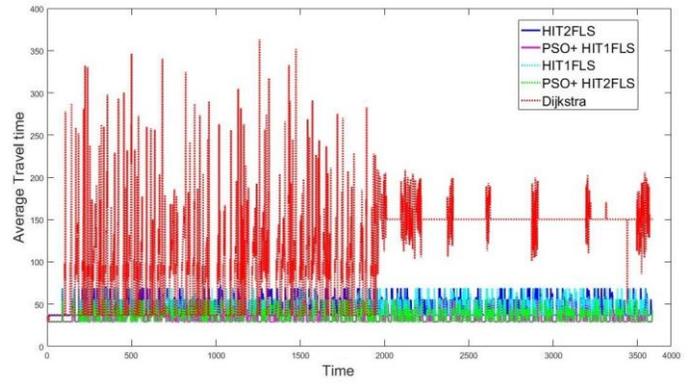

Fig. 15. Average travel time of vehicles (from origin to destination) (scenario3)

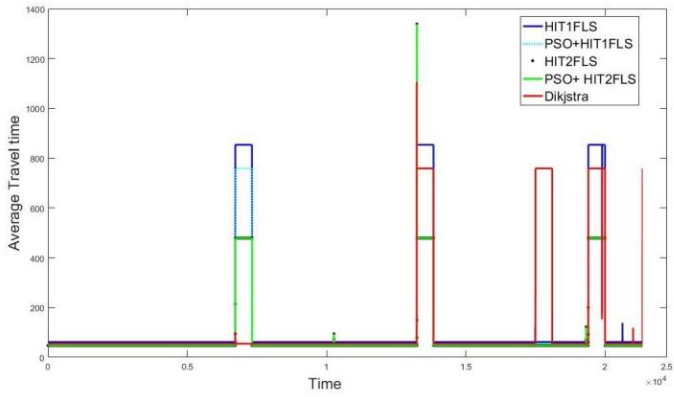

Fig. 13. Average travel time of cars (from origin to destination) (scenario 2)

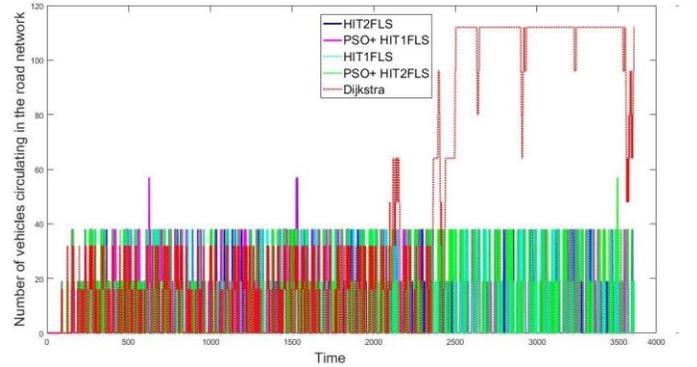

Fig. 16. Number of vehicles circulating in the road network (scenario3)

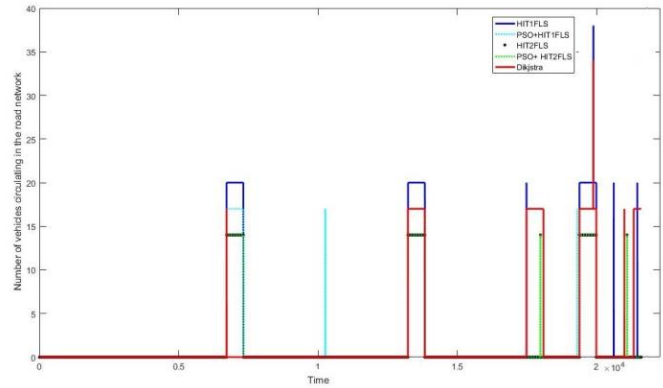

Fig. 14. Number of vehicles circulating in the road network (scenario 2)

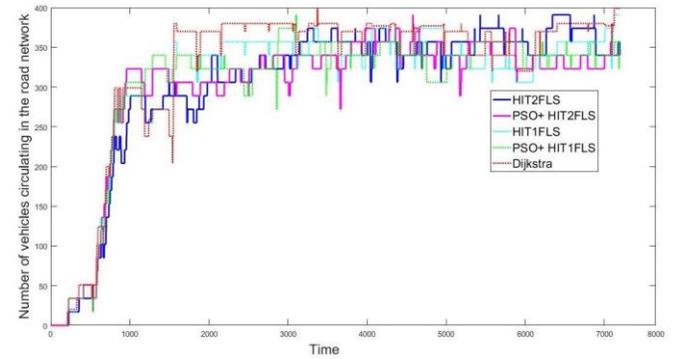

Fig. 17. Number of vehicles circulating in the road network (scenario4)

TABLE VII
MEAN OF AVERAGE TRAVEL TIME (SCENARIO 1)

|  | Mean of average travel time under 3600 sec |
|---|---|
| Dijkstra | 465.5000 |
| HIT1FLS | 193.4917 |
| PSO+HIT1FLS | 175.7889 |
| HIT2FLS | 146.6778 |
| PSO+HIT2FLS | 96.4056 |

TABLE VIII
MEAN OF AVERAGE TRAVEL TIME (SCENARIO 2)

|  | Mean of average travel time under 21600 sec |
|---|---|
| Dijkstra | 07.2565e+04 |
| HIT1FLS | 07.1120e+04 |
| PSO+HIT1FLS | 06.3292e+04 |
| HIT2FLS | 06.2537e+04 |
| PSO+HIT2FLS | 06.2537e+04 |

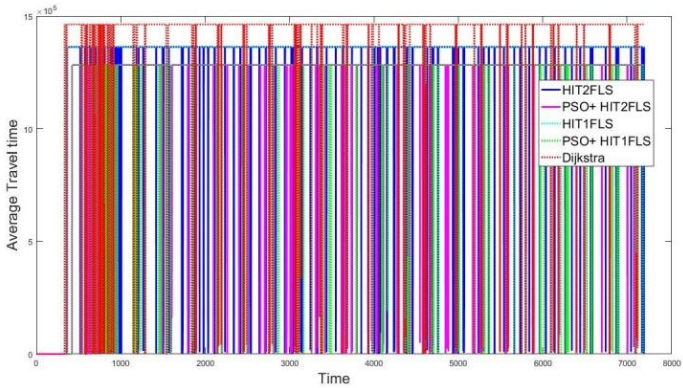

Fig. 18. Average travel time of vehicles (from origin to destination)(scenario4)

TABLE IX
MEAN OF AVERAGE TRAVEL TIME (SCENARIO 3)

|  | Mean of average travel time under 3600 sec |
|---|---|
| Dijkstra | 112.5167 |
| HIT1FLS | 44.1889 |
| PSO+HIT1FLS | 35.2944 |
| HIT2FLS | 44.3028 |
| PSO+HIT2FLS | 34.9556 |

Regarding scenario1, around 500 vehicles were encompassed in the simulation from various origins and destinations belonging to the same road network as shown in Figure 6.
The allowed speed for some roads in the center of the network (the interior of the network) is 20 m/s (meter/second) and other roads are considered as reasonably elevated speed roads with 40 m/s. In this context, a congestion situation was enforced by the addition of a high number of cars simultaneously in the nearest roads. In Figure 10, 100 vehicles (travel demands) were added in area A from 1000th to 1500th seconds.
As shown in Figures 11 and 12 and after simulation, the whole network is more adjustable in terms of average travel time and the number of vehicles in the network. Figure 11 reveals that Dijkstra algorithm fails to provide the shortest travel time due to congestion. With our approach, the travel time is assessed taking into account both distance and congestion. For a better comparison, this experiment considered in Figures 11 and 12 , we resume results in a quantitative way, by measuring the mean of average travel time under travel steps, given by the following Table 7. According to Table 10, the itinerary suggested to the same vehicle is based on two methods into the same road traffic simulation conditions, taking into account congested roads. It is noteworthy that the three possible paths have the same average speed, but the corresponding travel times are different and depend on the itinerary selection method. To show that our system behaves well with many realistic traffic scenarios, we decided to apply it with European cities like Luxembourg, Bologna and Cologne. The scenarios must be big enough to show the standard congestion patterns noticeable in modern cities. Figure 7 shows the topology of the LuST Scenario, with streets colored by type. The highway is depicted in blue, the main arterial roads in red and the residential roads in black. With respect to the quality of traffic flow in the road network, Figure 13 presents the advantages of the proposed EHIT2FLS, in the decrease of the average travel time in the entire road network during 6 h, compared to the Dijkstra's method, to the HIT1FKRS and to the HIT1FKRS optimized by PSO. In fact, the Table 8 confirms the global results illustrated by Figures 13 and 14. In this scenario, we have almost similar results with HIT2FLS (6.2537e+04) but it does not mean that it is better than other methods and that it offers a better global road traffic quality and avoids congested/jammed states.

TABLE X
MEAN OF AVERAGE TRAVEL TIME (SCENARIO 4)

|  | Mean of average travel time under 7201 sec |
|---|---|
| Dijkstra | 1.4731e+06 |
| HIT1FLS | 1.2261e+06 |
| PSO+HIT1FLS | 1.1444e+06 |
| HIT2FLS | 1.2124e+06 |
| PSO+HIT2FLS | 1.1372e+06 |

TABLE XI
ITINERARY SELECTION OF ONE VEHICLE (FROM ORIGIN ROAD 28 4 TO DESTINATION ROAD 29 4 )

| Intersection | Path selected using Dijkstras Algorithm | Path selected using EHIT2FLS | Other possible Path |
|---|---|---|---|
| 1 | 28 3 | 30 2 | 28 3 |
| 2 | 28 2 | 30 3 | 28 2 |
| 3 | 29 0 | 30 0 | 28 1 |
| 4 | 29 1 | 29 0 | 28 0 |
| 5 | 29 2 | 29 1 | 27 9 |
| 6 | 29 3 | 29 2 | 27 8 |
| 7 | 29 4 | 29 3 | 27 7 |
| 8 | - | 29 4 | 29 4 |
| Average speed (m/s) | 20 | 20 | 20 |
| Travel time (s) | 881 | 740 | 1224 |
| Distance (m) | 5023 | 9560 | 12350 |

For the third scenario, concerning the number of vehicles circulating in the road network, Figure 16 shows that EHIT2FLS helps a high number of vehicles reaching their destinations early, compared to other methods. Thus, Figure 15 bears out the effectiveness of our system in terms of average travel time, revealing that it is better than other methods. In Table 9, the mean average of travel time of EHIT2FLS is the minimum (34.9556) compared to other methods. In fact, these tables reveal the improvement of the results using PSO for learning our HIT2FKRS in terms of average travel time compared to the Dijkstra method, HIT2FKRS, HIT1FKRS and HIT1FKRS optimized by PSO. Therefore, about 40% of vehicles have altered their itineraries after the running of EHIT2FLS. The simulation results include the impact of the six abovementioned external factors: density, maximum allowed speed in the itinerary, familiarity of the driver with the roads, usual driving speed, departure time, and weather information. The results have affirmed the significant impact of the chosen contextual factors and the efficacy of the proposed EHIT2FLS is efficient and capable to ameliorate the quality of traffic network and circumvents congested/jammed states.

## VII. CONCLUSION

In this paper, an Evolutionary Hierarchical Interval Type-2 Fuzzy Knowledge Representation System for Travel Route Assignment was presented. This advanced traffic management system was developed based on Hierarchical Interval Type-2 Fuzzy Logic model optimized by the Particle Swarm Optimization (PSO) method. This allows the intelligent and prompt adjustment of the road traffic in the network according to the real-time changes. The itinerary selection is based on both traffic quality and itinerary length in the first stage and on a set of the most important contextual factors pertaining to the driver, the environment, and the path integrated in a hierarchical fuzzy system. On the one hand, the proposed algorithm contributes to an improvement in traffic fluency (the road network can support more vehicles without decreasing the average speed of vehicles) while taking into account the real-time road traffic information. On the other hand, it contributes in reducing the number of traffic congestion situations by avoiding the massive use of the same road at the same time (providing the suggestion of itineraries with a lower travel times). For the evaluation of our system, we used SUMO, a well-known microscopic traffic simulator. For this study we have tested four big realistic traffic scenarios located in the cities of Sfax, Luxembourg, Bologna and Cologne. Simulation results confirm that the proposed system (EHIT2FKRS) offers a better road traffic quality on the entire road network without a great loss on individual travel time compared to the Dijkstra method, HIT2FKRS, HIT1FKRS and HIT1FKRS optimized by PSO. Furthermore, our EHIT2FKRS can improve the number of vehicles that reach their destination and the mean journey time. As perspective, we intend in the near future to provide communications among nearby vehicles in order to know the state of network in a global way (accidents/Jams). Taking advantage of V2V communications for traffic management with the effectiveness of our EHIT2FLS, we can offer good road traffic quality of all regions of city.

## ACKNOWLEDGMENT

The research leading to these results has received funding from the Ministry of Higher Education and Scientific Research of Tunisia under the grant agreement number LR11ES48.